%% file: main.tex
\documentclass[letterpaper,twocolumn,10pt]{article}
\usepackage{usenix-2020-09}
\usepackage[available,functional,reproduced]{usenixbadges}

\usepackage{tikz}
\usepackage{amsmath}

\usepackage[utf8]{inputenc} % allow utf-8 input
\usepackage[T1]{fontenc}    % use 8-bit T1 fonts
\usepackage{hyperref}       % hyperlinks
\usepackage{url}            % simple URL typesetting
\usepackage{booktabs}       % professional-quality tables
\usepackage{amsfonts}       % blackboard math symbols
\usepackage{nicefrac}       % compact symbols for 1/2, etc.
\usepackage{microtype}      % microtypography
\usepackage{xcolor}         % colors
\usepackage{algorithm}
\usepackage{algpseudocode}
\usepackage{dsfont}
\usepackage{array,multirow}
\usepackage{graphicx}
\usepackage{subcaption}

\microtypecontext{spacing=nonfrench}

\usepackage{url}
\usepackage{xspace}
\usepackage{booktabs}
\usepackage{diagbox}
\usepackage{multirow}

\usepackage{enumitem}
\setlist{leftmargin=5mm}
\setlist[1]{itemsep=0pt}

\newlist{compactdesc}{description}{3}
\setlist[compactdesc]{topsep=0pt,partopsep=0pt,itemsep=0pt,parsep=0pt}

\usepackage{soul}
\sethlcolor{lightgray}

\input{macros}

\begin{document}

%don't want date printed
\date{}

% make title bold and 14 pt font (Latex default is non-bold, 16 pt)
\title{URET: Universal Robustness Evaluation Toolkit (for Evasion)}

\author{
{\rm Kevin Eykholt\thanks{These authors contributed equally.}}\\
IBM Research
\\ \\
 {\rm Jiyong Jang}\\
IBM Research
\and
{\rm Taesung Lee\textsuperscript{*}}\\
IBM Research
\\ \\
 {\rm Ian Molloy}\\
IBM Research
 \and
 {\rm Douglas Schales}\\
IBM Research
\\ \\
 {\rm Masha Zorin}\\
University of Cambridge
} % end author

\maketitle

\input{sections/abstract}

\input{sections/introduction}
\input{sections/related}
\input{sections/motivation}
\input{sections/design}
\input{sections/implementation}

\input{sections/experiments}
\input{sections/discussion}
\input{sections/conclusion}

\bibliographystyle{plain}
\bibliography{main}

\clearpage
\input{sections/appendix}

\end{document}

%% file: macros.tex
\newcommand{\eg}{{\it e.g., }}%
\newcommand{\ie}{{\it i.e., }}%
\newcommand{\etal}{{\it et al. }}%

\newcommand{\paratitle}[1]{\noindent\textbf{\textit{#1.}}\xspace}

\newcommand{\argmax}{\operatornamewithlimits{argmax}}

\def\sec{Sec}

\newcommand{\brute}[0]{Brute-Force\xspace}
\newcommand{\lookup}[0]{Lookup Table\xspace}
\newcommand{\external}[0]{Model Guided\xspace}

\newcommand{\beam}[0]{Beam Search\xspace}
\newcommand{\simanneal}[0]{Simulated Annealing\xspace}

\newif\ifsubmit
\submittrue

\ifsubmit
    \newcommand{\ke}[1]{}
    \newcommand{\tl}[1]{}
    \newcommand{\im}[1]{}
    
    \newcommand{\ds}[1]{}
    
    \newcommand{\todo}[1]{}
\else
    \newcommand{\ke}[1]{\textcolor{blue}{\{Kevin: #1\}}}
    \newcommand{\tl}[1]{\textcolor{orange}{\{Taesung: #1\}}}
    \newcommand{\im}[1]{\textcolor{magenta}{\{Ian: #1\}}}
    
    \newcommand{\ds}[1]{\textcolor{purple}{\{Doug: #1\}}}
    
    \newcommand{\todo}[1]{\textcolor{red}{TODO: #1}}
\fi

%% file: sections/abstract.tex
\begin{abstract}

Machine learning models are known to be vulnerable to adversarial evasion attacks as illustrated by image classification models. Thoroughly understanding such attacks is critical in order to ensure the safety and robustness of critical AI tasks. However, most evasion attacks are difficult to deploy against a majority of AI systems
because they have focused on image domain with only few constraints.
An image is composed of homogeneous, numerical, continuous, and independent features,
unlike many other input types to AI systems used in practice.
Furthermore, some input types include additional semantic and functional constraints that must be observed to generate realistic adversarial inputs.
In this work, we propose a new framework to enable the generation of adversarial inputs irrespective of the input type and task domain.
Given an input and a set of pre-defined input transformations, our framework discovers a sequence of transformations that result in a semantically correct and functional adversarial input. We demonstrate the generality of our approach on several diverse machine learning tasks with various input representations. We also show the importance of generating adversarial examples as they enable the deployment of mitigation techniques.

\end{abstract}

%% file: sections/introduction.tex
\section{Introduction} \label{sec:intro}

The powerful automation capabilities of AI have been adopted to empower and drive numerous data-driven tasks. However, the safety, security, and privacy of machine learning has become a concerning issue. In particular, adversarial machine learning, which studies how small perturbations on the input by active adversaries can predictably influence model behavior~\cite{szegedy2014intriguing}, indicates that use of AI in safety and security critical tasks such as cybersecurity may introduce new vulnerabilities into the system~\cite{cwe1039}.

To secure machine learning algorithms against evasion attacks, penetration testing of victim models is required to identify vulnerabilities and obtain a representative set of adversarial inputs. Recent studies in adversarial machine learning have developed numerous attack algorithms~\cite{papernot2016limitations, goodfellow2014explaining, Carlini2016TowardsET, athalye2018obfuscated, croce2020reliable, deepfool, zoo}. 

These approaches typically make small numerical changes to the input such as adding the loss gradient or averaging two inputs and examining the influence on the output of the model. However, many of these attacks are designed to target the image classification tasks, which limits their practical use for testing real systems. Image inputs are often a collection of continuous, numerical, and independent features. Furthermore, the preprocessing is often differentiable. These properties are key requirements for many existing attacks as they craft adversarial examples through backpropagation of the loss gradient through the classification pipeline, but many machine learning models use inputs lacking these properties. For example, in malware detection~\cite{2018arXiv180404637A, 10.1145/3433210.3453086, suciu:2019}, the raw input is a discrete object (\ie a binary file) and the corresponding feature representation may be a combination of discrete, continuous, and categorical values (\eg file length, library imports, file type), some which may not be independent. These features were necessarily obtained through non-differentiable feature extraction. As such, image-based adversarial machine learning attacks cannot be used to generate an adversarial binary nor an adversarial set of features. In many machine learning applications, the input properties are often only loosely fulfilled in the feature space as the classifier input often must be numerical.

Many defenses relied on a functional evasion attack for their development and deployment. Adversarial training, for example, trains a model on adversarial samples generated on-the-fly to mitigate the effect of evasion attacks~\cite{madry2018towards}. As the image-based evasion attack assumptions for non-image classifiers are often only generally true in the feature space, adversarial training currently is limited to training on adversarial features. However, the adversarial features used for training may not be representative of real inputs due to semantic correctness and functionality constraints, \ie there exists no input object with the generated adversarial features. Indeed, in Table~\ref{tab:intro:at}, we found adversarial training on adversarial features generated by an image-based evasion attack was ineffective against functional and semantically correct adversarial input objects generated using our framework.

\begin{table}[tb]
    \centering
    \caption{Effect of feature-level adversarial training. Standard training denotes training on the original unmodified dataset.} \label{tab:intro:at}
    \resizebox{\columnwidth}{!}{%
    \begin{tabular}{c|cc} 
        \toprule
         Training Method & Benign Acc. & Adversarial Acc. \\
        \midrule
        Standard & 92\% & 19\% \\
        Adversarial Training & 82\% & 29\%  \\
        \bottomrule
    \end{tabular}
    }
\end{table}

One common approach to enable evasion attacks for non-image domain tasks is to retool existing image-based evasion attacks to enforce additional domain specific constraints that ensure the semantic correctness of the generated adversarial features~\cite{grosse2017adversarial,Huang2018}. While this approach generates semantically correct adversarial features, it remains non-trivial to find an adversarial object with features. The issue remains that adversarial gradients cannot propagate through the non-differentiable mapping of input objects to their numerical feature representation. 

Another approach is to pick a task domain and define a domain specific set of \emph{functionality preserving input transformations} for the input object. These transformations define the basic manipulation operations used to craft adversarial examples~\cite{anderson2017, Demetrio2020AdversarialEA} during the attack process replacing the traditional addition and subtraction of adversarial noise to a real valued input. As transformations are performed directly on the object rather than the object's feature representation, the problem of non-differentiable feature extraction is bypassed. However, most works of this type make domain specific assumptions and use customized attack algorithms, making it difficult to easily adapt these works to other domains.

% \begin{figure}[t]
% \centering
% \includegraphics[width=0.5\linewidth]{figures/flowchart.pdf}
% \caption{Usage flowchart}\label{fig:ai}
% \end{figure}

To this end, a \emph{general purpose evasion attack framework} for AI systems and implement URET (Universal Robustness Evaluation Toolkit (for Evasion)), a toolkit that can be integrated into machine learning evaluation and remediation pipelines. We define a set of functionality and semantic preserving transformations for several common data representations. In addition, we also expose a transformation interface with which a user can define their own set of transformations for given data representation. These transformations establish the basic adversarial modification operations to be used by our framework. We then characterize the adversarial input generation process as a \emph{graph exploration problem}. Given an initial unperturbed sample, our framework finds a sequence of edges (\ie sequence of transformations) that achieve the adversarial objective (\ie misclassification) and generates an adversarial input. 

From the user's perspective, they simply need to identify the input data type(s) and how those types should be modified. They also select the graph exploration parameters, namely the vertex scoring function, the vertex ranking algorithm, and the graph search algorithm. Each of these components come pre-installed in the toolkit. Users can optionally define custom input constraints and inter-feature dependencies, which our framework automatically enforces during sample generation. Our contributions are as follows:
\begin{itemize}
    \item A general purpose adversarial input generation framework that formulates the process as a graph exploration problem in order to generate adversarial inputs irrespective of the input representation. Unlike prior works, which are limited in scope or focused on a specific domain, our framework can generate adversarial samples for any machine learning task irrespective of input data representation.
    \item As part of an ongoing adversarial evaluation effort, we document three case studies on machine learning tasks with different input types, highlighting the functionality and use cases of our framework.
    \item We demonstrate how our work can be used for adversarial remediation through integration with an adversarial training pipeline, a popular defense technique against evasion attacks.
    \item An open-source implementation of our framework available at \url{https://github.com/IBM/URET}.
\end{itemize}

%% file: sections/related.tex
\section{Related work} \label{sec:related}

\begin{table*}[!tb]
    \centering
    \caption{Comparison of supported functionalities of non-image evasion attack tools. $\bigcirc$: supported; $\bigtriangleup$: partially supported; $\times$: not supported. Typical tabular data includes categorical, boolean, integer, and float features. A config interface refers to the ability to quickly define and evaluate a machine learning model through use of a configuration file, command line call, or similar interface. All of the works here have implementations, but they either do not provide a user guide, are not maintained as the implementation exists for reproducibility, or are private.}
    \label{tab:support}
    \begin{tabular}{c|ccc|c|cc|c}
    \toprule
         \multirow{3}{*}{Attacks} & \multicolumn{3}{c|}{Input Types} & \multirow{3}{*}{Config Interface} & \multicolumn{2}{c|}{Opt. Goal} & \multirow{3}{*}{Open Source} \\
         \cmidrule{2-4} \cmidrule{6-7}
         & Tabular & Text & Custom & & Model & Feature &  \\ \midrule
        SLEIPNIR \cite{Huang2018} & $\times$ & $\times$ & Malware & $\times$ & $\bigcirc$ & $\times$ & $\bigtriangleup$ \\
        Gym-Malware \cite{anderson2017} & $\times$ & $\times$ & Malware & $\times$ & $\bigcirc$ & $\times$ & $\bigtriangleup$\\
        \midrule
        Graph Search \cite{KulynychHST18} & $\bigtriangleup$ & $\times$ & $\times$ & $\times$ & $\bigcirc$ & $\bigcirc$ & $\bigtriangleup$ \\
        Pierazzi \etal \cite{pierazzi2020problemspace} & $\bigcirc$ & $\bigcirc$ & $\bigcirc$ & Unknown & $\bigcirc$ & $\times$ & $\bigtriangleup$ \\
        Counterfit \cite{counterfit} & $\times$ & $\bigcirc$ & $\times$ & $\bigcirc$ & $\bigcirc$ & $\bigcirc$ & $\bigcirc$ \\ \midrule
        URET (Ours) & $\bigcirc$ & $\bigcirc$ & $\bigcirc$ & $\bigcirc$  & $\bigcirc$ & $\bigcirc$ & $\bigcirc$  \\
    \bottomrule
    \end{tabular}
\end{table*}

Although adversarial machine learning is well-studied, most of its development has been focused on image-related tasks, such as object classification, image segmentation, and instance segmentation. However, powerful automation capabilities provided by machine learning has encouraged its use in other domains~\cite{Rudd2019}. Non-image task domains such as cybersecurity, text classification, and even traditional tabular datasets pose issues for adversarial machine learning as more considerations must be made when manipulating input objects. Unlike images, most other input objects (\eg malicious binaries, text strings, tabular data) must also remain semantically correct and operational after adversarial manipulation. Here, we discuss related works that make advances towards developing adversarial attacks for non-image data representations.
We summarize some of more generic and promising frameworks in Table~\ref{tab:support}, and compare their functionalities.

\subsection{Domain Specific Attacks}
One group of related works focuses on designing adversarial attacks with domain specific assumptions. Given its security critical nature, many recent works focus on the malware classification task. Grosse \etal refactored the Jacobian Saliency Map attack (JSMA) to enable manipulations of malware binaries~ \cite{grosse2017adversarial}.  Given a set of binary features (\ie $[0,1]$ features) for a malware sample, where a value of $1$ indicates that the sample exhibited the feature (\eg the malware imported a specific library).  With such a feature representation, their proposed attack computes the Jacobian Saliency map of the sample and finds the feature indices that would most likely cause adversarial misclassification. As they purposely used an interpretable feature representation, adversarial modifications were easily reproducible. Malware functionality was also preserved as their attack only modified 0-valued features. That is to say, the adversarial attack adds new capabilities rather than removes it. This approach using limited manipulations of binary feature representations has been leveraged by other works as well~\cite{Huang2018}. The limitation of works that refactor existing image-based adversarial attacks is a requirement for interpretable, independent feature representations in order to easily map modifications back to the original input representation.

Other works have forgone adapting image-based attacks and proposed their own custom algorithms that generate adversarial cybersecurity objects~\cite{anderson2017, Demetrio2020AdversarialEA, 10.1145/3433210.3453086}. They first define a set of functionality preserving input object transformations and use varying methods to find a sequence of transformations to cause misclassification. Demetrio \etal and Lucas \etal used the traditional approach of optimizing an objective function to find the sequence of adversarial modifications \cite{Demetrio2020AdversarialEA, 10.1145/3433210.3453086}. Anderson \etal trained a reinforcement learning agent to solve this problem \cite{anderson2017}. Although these works share similarities with our framework, they are limited to support a single data representation and a task. On the contrary, our framework is \textit{easily customizable for any data representation} through modification of a configuration file and, possibly, the transformation interface.

\subsection{Comparison with generic frameworks}

Of interest are works that, like ours, propose a generic adversarial generation framework with an open-source implementation. Kulynych \etal \cite{KulynychHST18} propose representing the process of generating adversarial samples as a graph search problem. Similar to our framework, their methodology seeks to find a sequence of transformations that cause misclassification, but we note a few key differences. First, they focus their framework on binary classifiers given their security relevance. Although the case studies we show later are also binary classification tasks, our exploration objectives support non-binary classification tasks as well. A second difference is how edges and nodes are evaluated and explored during graph exploration. They focus on the A* search algorithm as it will find an optimal adversarial example, under certain assumptions, measured by the fewest number of transformations. The current version of our framework implements several exploration configurations recognizing that a user's needs may vary. For example, some of our exploration configurations implement predictive analytics, something their framework does not support, which trade adversarial success rate for lower runtime. This trade-off can be useful for exploring large graphs or if speed is a concern, \eg adversarial training. They attempt to alleviate this issue using preset heuristics, but the effect of this approach is unclear. Of issue is their proposed preset heuristics are based on the $p$-norm, which is not ideal to compare categorical features. This limits their approach to numeric types such as integer and Boolean features. Finally, the authors have only provided code and instructions to reproduce their experiment results\footnote{https://github.com/spring-epfl/trickster}, but do not appear to have a plan to support a set of general purpose tools and guidelines for use by the larger community, which is an issue our work is trying to address.

A closely related work is that of Pierazzi \etal who formalized ``problem-space'' evasion attacks and proposed a general attack framework~\cite{pierazzi2020problemspace}. They also observe extracting feature representations of inputs in the problem-space (\ie what we denote as objects in the paper) is a non-invertible and non-differentiable process and hinders traditional evasion attacks. From our reading of the paper, their approach requires the user to define a set of domain specific input transformations and constraints. Then, their framework uses either a problem driven (\eg genetic algorithms, Monte Carlo search tree, etc.), gradient driven, or hybrid search algorithm to discover a sequence of transformations that result in misclassification. In their paper, they focus on Android malware classification as an example task, but describe multiple other tasks in Table I of their paper. Unfortunately, due to ethical concerns, their implementation is private\footnote{They only allow academic researchers access to their code.}, which prevents us from comparing their implementation with ours and understanding the user interface or operation. However, it remains true that their tools are not intended to be integrated into model evaluation and remediation pipeline in its current state.

Recently, Microsoft released Counterfit, an open source toolkit for testing AI systems against adversarial machine learning attack \cite{counterfit}. Despite rising concerns regarding the vulnerability of AI systems to adversarial attacks, most of the businesses they surveyed were not prepared nor had the tools necessary to secure AI systems \cite{microsoft_survey}. Although several libraries exist for deploying machine learning attack, most of these libraries are designed for image and text inputs~\cite{art, papernot2018cleverhans, morris2020textattack, goodman2020advbox, ding2019advertorch}. To fill the need for a general adversarial evaluation toolkit, Counterfit builds upon three existing libraries, the Adversarial Robustness Toolbox (ART)~\cite{art}, TextAttack~\cite{morris2020textattack} and AugLy~\cite{papakipos2022augly}, and exposes a command line interface to run adversarial attacks and evaluations from these frameworks. Counterfit serves to lessen the burden of knowledge for users when attempting to deploy adversarial attacks against their own systems. Thanks to the use of blackbox adversarial attacks from ART and attacks from the TextAttack library, Counterfit can support more input types and task domains than traditional adversarial attacks. As a framework, Counterfit helps to simplify the adversarial attack and evaluation process for security experts by abstracting away the details such as hyperparameter tuning and attack selection.

Although Counterfit is a major step towards enabling adversarial evasion attacks and evaluations in a general context, it is limited by the attack libraries it uses in the background. First, Counterfit cannot support uncommon input objects such as file binaries because the attacks it employs operate in the feature space. Like the work by Grosse \etal and Huang \etal~\cite{grosse2017adversarial, Huang2018}, Counterfit can only be used to generate an adversarial input object if the features are interpretable. Second, Counterfit cannot enforce custom input constraints and inter-feature dependencies, as such constraints are not supported in ART or TextAttack.

%% file: sections/motivation.tex
% \section{Motivation}
\section{Preliminary}
\label{sec:motivation}

\begin{figure}
\centering
\includegraphics[width=0.7\linewidth]{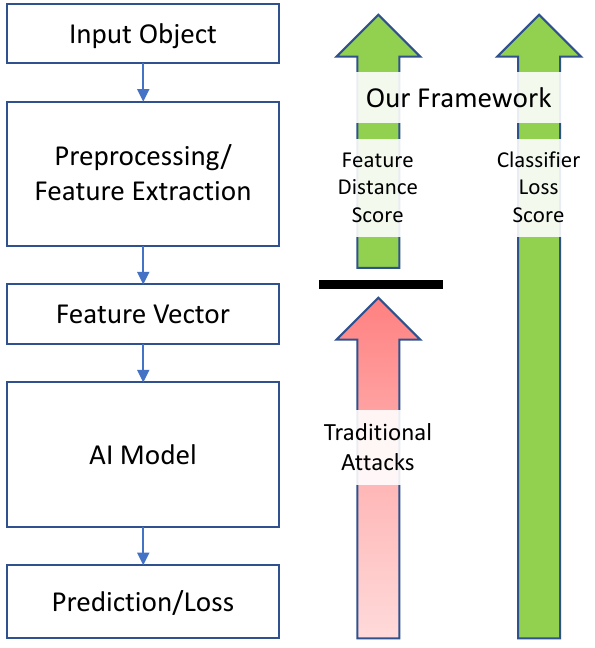}
\caption{A typical model pipeline. Traditional adversarial attacks, which use loss gradients, are limited to generate adversarial feature vectors. Our framework can generate adversarial input objects or adversarial features using the classifier loss scoring function. Furthermore, our framework remains backward compatible with existing attacks using the feature distance scoring function. }\label{fig:ai}
\end{figure}

In this section, we describe the challenges and need for a general adversarial attack and robustness evaluation framework.
To avoid confusion, for the remainder of the paper we will use the following terminology to describe the data going through an AI model (Figure~\ref{fig:ai}):
\begin{itemize}
    \item \emph{Object:} An \textbf{object} denotes the input value to the AI system before feature extraction has occurred. An object can be a singular data type or a collection of data types. A data type could be an abstract data type (\eg domain name or binary) or a common data type (\eg numerical, categorical, and textual data).
    \item \emph{Feature:} A \textbf{feature} denotes the input value to the machine learning model after feature extraction has been performed, and it is a numerical value such as a real number or integer. A \textbf{feature vector} is composed of one or more features. We may use features and feature vector interchangeably.
    \item \emph{Input:} An \textbf{input} denotes a value anywhere before the model's input layer (\ie either an object or a feature vector).
\end{itemize}

Using the above terminology, a typical AI system consists of a feature extractor and a machine learning model.
Given an \emph{object} of a known structure, the system first converts the object into a numerical representation recognized by the machine learning model, optionally normalizing the feature values. This \emph{feature vector} is processed by the machine learning model and an output prediction is made.

\subsection{Adversarial Machine Learning}

The increasing use of AI in safety and security critical systems has heightened concerns regarding the vulnerabilities AI introduces to such systems. Prior work with image-based classifiers has identified one class of attacks called \textit{adversarial evasion attacks}~\cite{szegedy2014intriguing, goodfellow2014explaining}. In such attacks, an adversary computes a precise set of input manipulations such that the perturbed input is misclassified. Such attacks threaten the reliability of AI systems as adversaries with sufficient access can control the output decisions of the classifier. Many have proposed solutions to mitigate the effect of adversarial evasion attacks, but few have succeeded ~\cite{athalye2018obfuscated, papernot2016distillation, madry2018towards, wong2018provable,cohen2019certified,zhang2019stable}. The main issue early adversarial defenses suffered was a reliance on gradient obfuscation, a technique that prevents correct estimation of a classifier's loss gradient, usually due to the use of non-differentiable operations. As early adversarial evasion attacks relied on the classifier's loss gradient to guide the adversarial image perturbations, gradient obfuscation based defenses appeared very effective at preventing adversarial evasion attacks. Unfortunately, Athalye~\etal demonstrated that gradient obfuscation can be easily bypassed through use of alternative gradient approximation methods~\cite{athalye2018obfuscated}. Despite the failure of early defenses, there exist several training-based defenses that are effective at improving the robustness of image-based classifiers against adversarial evasion attacks~\cite{madry2018towards, wong2018provable,cohen2019certified,zhang2019stable}.

\subsection{Challenges}

An evasion attack can be formulated as finding $T(x)$ for an input $x$ such that
\begin{align}\begin{split}
    F(T(x)) \neq F(x) & \quad\mathrm{if\ untargeted\ attack,}\\
    \argmax F(T(x)) = c(x) & \quad\mathrm{if\ targeted\ attack.}
\end{split}\end{align}
where $F$ is an AI model, $c(x)$ is the target class, and $T$ is a function applying a small change.
For an image classification task, it is typically $T(x) = x+\delta$ such that $\|\delta \|_p < \epsilon$, $\| \cdot \|_p$ is $p$-norm, and oftentimes $\delta$ is derived from the gradient of $F$ with respect to $x$ or through other numerical operations on $x$. However, how one can define $T$ becomes nontrivial as we step out of the image domain.

% As adversarial machine learning attacks have been traditionally studied in the context of image recognition, it is non-trivial to apply them elsewhere.
Unlike the image domain where we can easily apply numerical modifications to $x$, doing so in other domains requires addressing the following challenges: 1) non-differentiable feature extraction;  2) variable input types; and 3) input semantics and functionality preservation.

\paratitle{Non-Differentiable Feature Extraction}
Most state-of-the-art adversarial evasion attacks generate adversarial inputs by computing the loss gradient with respect to the model's input and using the loss gradient to guide the adversarial modifications. Within the model, computing the loss gradient is straightforward as machine learning models are composed of differentiable layers necessary for backpropagation. The challenge arrives when the loss gradient must be back propagated through the feature extraction layer. In the image domain, there is little to no structural difference between the object (the original image) and its extracted features as both are arrays of continuous, numerical values. When input transformations are performed, they are often done for data augmentation purposes and are differentiable (\eg shift and rotate). However, most other input objects are non-numerical, thus input transformations are required to obtain a numerical feature representation that the machine learning model can process. The difference in structure requires such transformations to be non-differentiable. As such the loss gradient cannot be backpropagated back to the input object preventing traditional attacks from generating adversarial objects. We are aware of gradient approximation techniques such as the Backward Pass Differentiable Approximation proposed by Athalye \etal \cite{athalye2018obfuscated}, but such techniques require that the differentiable approximation be similar to the original feature extraction function and only permit slight gradient inaccuracies. Thus, existing attacks are limited to generating adversarial features.

\paratitle{Variable Input Types} In image recognition, the input object (an image) consists of an array of numerical, continuous values each sharing the same data type. In most domains, however, the input data types are variable. In malware detection, the input object may be a file binary. In Domain Generation Algorithm (DGA) detection, the input sample may be a string representing the domain name. In classification tasks with tabular datasets, the input object may be an array of mixed data types (\eg numerical, categorical, text, etc). As the task changes, the structure and data type(s) of the input object changes as well. This variability also extends to the input feature vector as the feature vector may consist of a mix of discrete and continuous numerical values that represent numerical quantities or preprocessed categorical features. Traditional adversarial attacks are not equipped to handle such varied input representations.

\paratitle{Input Semantics and Functionality Preservation} Images and their respective feature vector representations are simple inputs to adversarially modify because the modifications is a simple incremental addition of noise. Furthermore, the inputs usually consist of independent features, \ie the value of one pixel is not affected by the value of other pixels in the image. Beyond clipping the input to ensure the features remain in a certain valid range, these properties allow existing adversarial attacks to largely ignore the need to preserve the input semantics or functionality of an image object. When considering adversarial modifications on general object types or feature representations, we must take these constraints into account. Otherwise, the generated adversarial inputs might be meaningless as they would represent impossible inputs or would change the original intent of the input. For example, when designing MalGAN, the authors only allowed the attack to add new features because if a certain feature were removed, the corresponding change to the malicious binary may prevent its intended operation~\cite{hu2017generating}. Other works looking to modify file binary rather than its feature representation also used custom modification functions to ensure the binary remained valid after modification~\cite{anderson2017, Demetrio2020AdversarialEA, 10.1145/3433210.3453086}.

%For example, the feature vector representation of a string object may contain features that counts the number of vowels, the number of consonants, and the total number of letters. If an adversarial attack seeks to increase feature counting the number of vowels, then either the feature counting the number of consonants must decreases or the feature counting the number of letters must increase. Otherwise, an illegal feature representation would be generated breaking input semantics. 

\subsection{Is adversarial machine learning a threat?}
% Preliminary studies in adversarial machine learning on computer vision systems assumed strong threat models in which the adversary had full visibility into the AI system and the ability to manipulate the input at any point before it was provided to the model. Eykholt \etal suggested that such a threat model was unrealistic in practice as an adversary with such high level access to the AI system could perform more damaging actions than just subtly manipulating model inputs~\cite{eykholt2018robust}. Furthermore, gaining such high privileged access would be difficult. In a weaker threat model in which the adversary can only manipulate the operational environment of an AI computer vision system, they demonstrated that there are additional environmental factors, which harden the task of adversarial evasion and prevent the success of existing adversarial attacks.
% However, once these environmental factors are taken into account as part of the attack pipeline, the adversarial attacks are successful once again. 
% Since the criticism on the unrealistically strong threat model
% in which the adversary had full visibility into the AI system and the ability to manipulate the input at any point before the model~\cite{eykholt2018robust},
Recently,
the research community has explored adversarial attacks to handle more diverse environmental conditionals and task domains against real world applications~\cite{DBLP:journals/corr/abs-1812-05271, Chen2018RobustPA, DBLP:journals/corr/abs-1803-04683} further fueling concerns about the security of AI systems.
NIST has been working on AI Risk Management Framework to foster the development of technologies to improve trustworthiness of AI including reliability, robustness, safety, and security~\cite{nist}.
In cybersecurity, researchers and industry have begun to recognize the threat of adversarial machine learning.
MITRE recently released Adversarial Threat Landscape for Artificial-Intelligence Systems (ATLAS)~\cite{mitre:atlas} to systematically describe common attack tactics and techniques used in adversarial machine learning similar to MITRE ATT\&CK framework~\cite{mitre:attack}, such as poisoning training data, evading ML model, and crafting adversarial data, with case studies. For example, a security vulnerability in the Proofpoint email protection system was discovered by building a copy-cat email classification model and evading the detection by crafting an adversarial spam email~\cite{proofpoint:evasion}.

To motivate the need for generic tools that enable the study of adversarial machine learning in safety and security critical tasks, we studied the robustness of state-of-the-art malware detectors. We used an open source malware detector and a set of functionality-preserving transformations to disguise malware programs as goodware to the detector. Then, we uploaded these samples to see if they are correctly classified by the 60+ antivirus products used on VirusTotal which are commonly used to label malware samples. 
% We use VirusTotal here as it is commonly used in industry to label malware samples. 
As Table~\ref{tab:motiv:ember} shows, the detection rate of the scanners on VirusTotal drops significantly, and some detectors could correctly detect only 6\% of malware programs we generated. These results shows that we can adversarially modify an input object that remains functional and transfer the modified object to other classifiers to evade detection. Thus, there is a strong need to be able to synthetically generate such variants before an adversary exploits this vulnerability so we can fortify classifiers accordingly.

\begin{table}[htb]
\centering
\caption{The detection rate per scanner per malware sample.}
\label{tab:motiv:ember}
\resizebox{.7\columnwidth}{!}{%
\begin{tabular}{c|ccccc} \toprule
    Dataset & Average & Min  & Max \\ \midrule
    Original & 82\% & 50\% & 90\%  \\
    % Adversarial (Random) & 72\% & 36\% & 88\%  \\
    Adversarial & 58\% & 6\% & 91\%  \\
    \bottomrule
\end{tabular}
}
\vspace{-5mm}
\end{table}

%% file: sections/design.tex
\section{Design}
\label{sec:design}

In this section, we describe our generic attack framework enabling the generation of adversarial inputs irrespective of the input type and task domain. We characterize the adversarial generation process as a \emph{graph exploration} problem in which we seek a sequence of edges from the original input vertex that results in a new vertex, but a different model prediction.

% Components overview

Our framework can be abstracted into two main steps: 1) Graph Definition (\sec~\ref{sec:graphdef}) and 2) Graph Exploration (\sec~\ref{sec:graphsearch}). First, the user defines the exploration task by specifying the set of input transformations to be used as well as any functional or semantically correct constraints (\sec~\ref{sec:constraints}). Next, the user defines the graph explorer by selecting a \textbf{vertex scoring function}, an \textbf{edge ranking algorithm}, and a \textbf{graph search algorithm} from our framework. With the input transformation and explorer defined, our framework will automatically explore the graph until the objective is achieved (\eg an adversarial input is discovered) or the exploration budget is exhausted (\eg number of transformations).

\subsection{Step 1: Graph Definition} \label{sec:graphdef}

\begin{figure}[t] \centering
\includegraphics[width=0.9\linewidth]{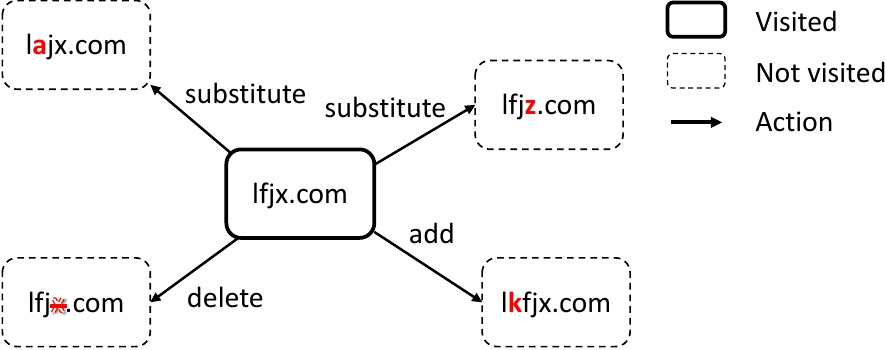}
\caption{An illustration of graph exploration.} \label{fig:graph}
\end{figure}

Our framework can be viewed as defining a graph $G=(V,E)$ where the set of vertices $V$ is all possible inputs and $E$ are the edge types\footnote{Note that this is a theoretical representation of the dynamic graph to be explored as building a static graph of all possible inputs is prohibitively expensive (\eg for domain names, there are up to $\sum_{i=3}^{63}36\times37^{i-1}$ vertices)}. The edge types $E$ are a set of semantic and functionality preserving input transformations that are chosen by the user. In our design, we only require that the input state after transformation remains semantically correct and functional. Therefore, their definition is flexible and can represent a variety of object modifications such as
changing a categorical value (\eg browser \texttt{Chrome/74.0.3729.169} to  \texttt{Safari/604.1}), 
manipulating a textual value (\eg domain name \texttt{lfjx.com} to \texttt{lfjz.com}),
or directly modifying an executable file. Our current implementation includes pre-defined transformers for some common basic data types (\ie numerical, categorical, text). Figure~\ref{fig:graph} shows an example for DGA detection where the input is represented by a string value with three edge types representing the different possible input transformations. Through our transformation interface, users can easily define their own transformers in cases where the input data type or transformation is currently not supported. We used this interface to also define a set of binary file transformations used in prior work~\cite{anderson2017}.

\subsection{Step 2: Graph Exploration} \label{sec:graphsearch}

\begin{figure}[t]
\centering
\includegraphics[width=0.6\linewidth]{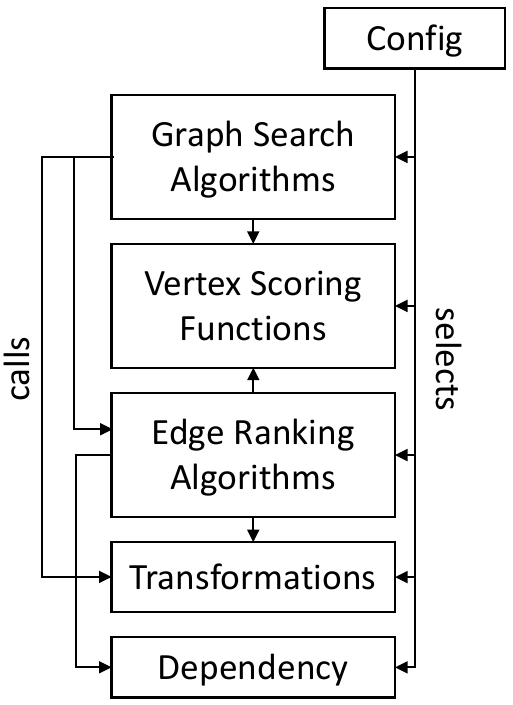}
\caption{Our framework structure. A configuration file is read by our framework and automatically constructs an graph explorer consisting of several components.}\label{fig:components}
\end{figure}

Once the graph has been defined through selection of the input data type and its corresponding transformations, the next step is to define a graph explorer. Starting at a vertex representing the current input, the explorer evaluates the connected edges and selects one or more edges based on the search method. The transformation corresponding to the selected edge(s) is applied and it is checked to see if the adversarial object has been achieved (\ie an adversarial input is found). If not, the edge is scored and exploration continues. This process is repeated with the selected vertices until the exploration objective is satisfied  or the exploration budget is exhausted. If exploration was not terminated early due to the exploration objective being achieved, then the ``best'' scoring vertex is returned. In Figure \ref{fig:components}, we show an overview of the graph explorer created by a configuration file.

\subsubsection{Vertex Scoring Function}
Our framework supports two vertex scoring functions by default: 1) classifier loss and 2) feature distance. The scoring function informs our framework  the ``closeness'' a vertex is achieving the objective. In general, the goal of our framework is to find a vertex that is adversarial, \ie the vertex represents an input different from the original input, is within some number of hops away from the original vertex (or within some norm bound of the original input), and is misclassified by the classifier. When found, exploration is terminated. The vertex scoring functions define a metric to guide exploration in achieving this objective. 

The \textbf{classifier loss scoring function} is used when the objective is to maximize the classifier loss for the current input $z$ as is done in traditional adversarial machine learning. Given a classifier $F$ and the classifier's prediction on the original input $y$, we score the vertex belonging to $z$ as follows:

\begin{equation}
    \label{eq:classifier_objective}
    score = \mathbf{L}(F(z), y)
\end{equation}
where $L$ is the classifier's loss function. Traditionally, the cross entropy function is used. If the user desires to generate an adversarial input that is misclassified as a specific target label $y_t$ rather than a random label that is not y, we modify Equation \ref{eq:classifier_objective} to minimize the loss with respect to the target label.

%\begin{equation}
%    \label{eq:classifier_objective2}
%    score = - \mathbf{L}(F(z), y_t)
%\end{equation}

While our framework can generate adversarial examples on its own, we recognize that some users would prefer to leveraging existing adversarial evasion attacks. With such attacks, a user would first generate an adversarial feature vector as they cannot propagate the modifications through the feature extraction component~\cite{papernot2016limitations, goodfellow2014explaining, Carlini2016TowardsET}. Then, the generated adversarial feature vector would be used by our framework as the exploration objective. The advantage of this approach is that existing adversarial attacks are well studied for numerical input types and may provide better results than relying on the classifier loss in certain scenarios. The \textbf{feature distance scoring function} encourages finding an input $z$ that when passed through feature extractor $E$ is as close to the target features $f$ as possible with respect to a distance function $D$. The vertex score is expressed as follows:

 \begin{equation}
    \label{eq:distance_objective}
    score = - D(E(z),f)
\end{equation}

\noindent where $E(z)$ is the feature representation of $z$ if $z$ is an object. If $z$ is already a feature vector, then $E$ is the identity function. The negative sign ensures that closer inputs with respect to the target features are given higher scores. As the framework may be unable to exactly find an input $z$ with input features $f$ due to semantic correctness and functionality constraints,  early termination with this scoring function is still possible if the current input is adversarial.

As a final point, although our current design pre-defines these two loss vertex scoring functions, our exploration framework can support other scoring methods if defined by the user. For example, the cost-based scoring function proposed by Kulynych \etal \cite{KulynychHST18} or a combination of multiple scoring functions could be used in our framework.

\subsubsection{Edge Ranking Algorithm}
%\paratitle{Exploration Algorithms}
Our framework pre-defines four edge ranking algorithms: a) Random, b) \brute; c) \lookup; and d) \external.

Given a vertex, the Random ranking algorithm randomly returns one or more edges without scoring them. It is the fastest of the ranking algorithms, but has the lowest success rate. As we will discuss later, this algorithm is usually used in combination with the \simanneal exploration algorithm.

Given a vertex, the \textbf{\brute} algorithm loops through each connected edge and applies the transformation specified by the edge (\ie visits the connected vertex). The score of a connected edge is equal to the vertex score of its connected vertex. For example, in Figure \ref{fig:graph}, assume the current vertex is ``lfjx.com'' and cross-entropy loss is used to score edges. For brevity, we'll also assume that the four edges shown are the only possible transformations. \brute would perform each transformation individually, and then classify the transformed inputs. Thus, it would generate the strings ``lajx.com'', ``lfjz.com'', ``lfj.com'', ``lkfjx.com'', score each one's edge based on its respective classification loss. As the transformations are done during runtime, the Brute-Force approach is often the slowest of the ranking algorithms. However, it is the most accurate as the edge rankings are based on the real state of the explored vertices.

The \textbf{\lookup} algorithm first runs a training phase using a small set of samples. For each training sample, its 1-hop neighborhood is explored. For each edge encountered, the current edge weight is computed using the Brute-Force approach and is stored a the lookup table, averaged with prior computed values of the edge weight. By exploring the 1-hop neighborhood of every training sample, the \lookup ranking algorithm generates a table of estimated edge weights for every unique edge encountered during training. During exploration, the ranking algorithm consults the table for the average edge weight of each connected edge and uses those values to rank the edges. As it does not consider the current vertex state, the edge ranking produced may differ from the true ranking, which often results in a lower success rate compared to Brute-Force. However, the table lookup is often faster than transforming the input as \brute does.

The \textbf{\external} algorithm also ranks edges without visiting the vertices by relying on a pre-trained model for edge selection.
Compared to \lookup algorithm, this provides the user a possibility to consider the current vertex state to predict the edge ranking more accurately.
A model can take the current vertex
and the vertex score to predict which edges are likely to improve the score.
While our framework can take any such model to rank edges, in our experiment,
we leverage Keras-RL~\cite{plappert2016kerasrl} to train a reinforcement learning policy as reinforcement learning considers both the current and future rewards for a particular transformation. As such, our framework includes support for implementing reinforcement learning policies to be used with the \external algorithm. We describe the implementation detail in the Appendix.
% Our framework does not provide pre-defined training functions to create such models, so it is expected that the user defines the training function or has pre-trained the model external to our framework. Once provided, our framework simply provides the current input state to it and expects a ranked list of connected edges to be returned.
% In our experiments, we tested this algorithm using a reinforcement learning model, who details can be found in the Appendix.

\subsubsection{Graph Search Algorithm}
Our framework implements two search algorithms by default: \beam and \simanneal. The search algorithm dictates which neighboring vertices are propagated to the next epoch of exploration.

The \textbf{\beam} algorithm is an algorithm that only passes the top-k best scoring nodes to the next epoch of exploration. The number of edges passed is determined by the \textbf{beam width}. We also include \textbf{depth} parameter, which defines the maximum number of edges to explore with respect to the original input and implicitly defines the \textbf{transformation budget}. The beam search algorithm can be combined with any of the edge ranking algorithms.

The \textbf{\simanneal} algorithm is a temperature-guided time-restricted random search algorithm. Unlike \beam, this algorithm can only be used with the Random ranking algorithm by definition and has an additional \textbf{time budget} parameter, which defines the per-sample exploration time. During each exploration epoch, it randomly selects a random length sequence of edges starting from the current input vertex, evaluates the new input state, and then determines if the new input state should be kept (i.e set as the new current state) based on the new vertex's objective score and the current \textit{temperature}. The temperature is a parameter that initially encourages exploration of new input states in the early phases of exploration. As time passes, the temperature gradually decreases, which discourages exploration and instead causes the algorithm to prefer to keep input states that better satisfy the exploration objective.

\subsection{Semantic Correctness and Functionality}
\label{sec:constraints}

A key component of our framework is to ensure the semantic correctness and functionality of generated adversarial inputs. Our framework defines modification functions, which we denote as \textbf{input transformers}, for a few basic input types: integers, floats, booleans, strings, categorical values, and one-hot categorical values. These input types are often used in most AI systems, at least at the feature representation level, and they can be combined to obtain new input representations. Furthermore, for most of these input types, the constraints on semantic correctness and functionality are usually very simple to ensure. For example, given a vector of floats, e.g., representing an image, the constraints take the form of ensuring the floats are bounded in a certain range.

Due to the wide adoption of machine learning, there are many possible task domains and a variety of input types. Prior works in malware detection~\cite{anderson2017, Demetrio2020AdversarialEA, 10.1145/3433210.3453086} used custom modification functions tailored for the task domain and our framework expands on this approach. Rather than enumerating every possible input type, we allow users to define new input transformations through an easy-to-use interface as users are likely to possess the necessary domain knowledge and understanding of input constraints. Through our interface, the user simply needs to define the transformation operation and, if necessary, methods to compute a set of possible transformations given an input value. Once defined, our framework automatically uses the transformations when encountering the respective input type. As the user provides the definition, the adversarial inputs generated by the framework will be semantically correct according to their definition. 

Additionally, our framework also includes the ability for users to define individual constraints on an input value or, in the case of a vector with multiple input values, dependency constraints between input values. The $L_p$ norm constraint used for image-based adversarial attacks is an example of an individual constraint for a float type of an input value. Another example is an edit distance for non-numerical inputs, such as strings. As inputs can have multiple representations, the definition of input transformations includes a function to enforce such constraints if necessary. After transforming an input, our framework checks if the user defines any dependency constraints as part of the attack. A dependency constraint is expressed as a function and multiple dependency constraints can be passed to our framework. For example, suppose that we have a feature vector of length three and the third index is computed by summing the first two indices. A user can express this relationship with a dependency function that takes in the current input state and the first two indices as input, sums their values, assigns it to the third index, and then returns the new input. 

%% file: sections/implementation.tex
\section{Using Our Framework}
\label{sec:implementation}
\begin{figure}[h] \centering
\includegraphics[width=\linewidth]{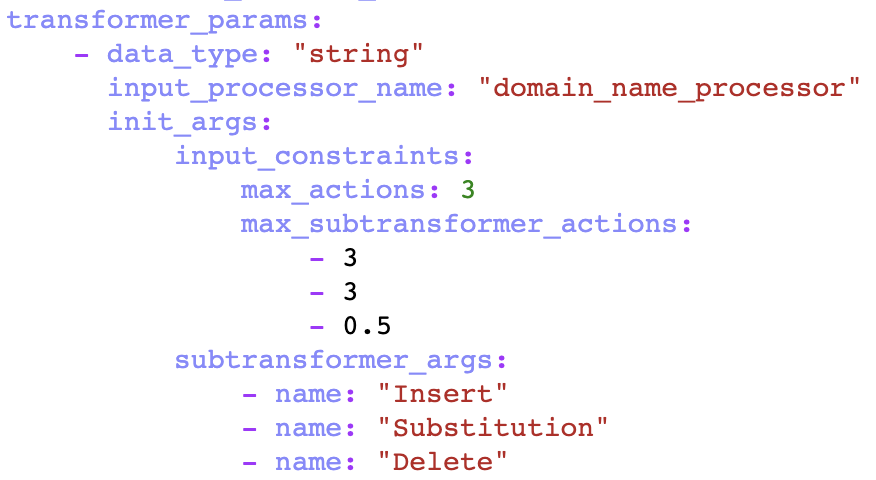}
\caption{The string transformer definition for the DGA task.} \label{fig:transformer_params}
\end{figure}

In this section, we'll give a brief overview of the typical user experience when using our framework. For this discussion, we will use the DGA classification task and the \beam (\brute) exploration configuration as an example. First, the user must define the set of transformations to be used. A user defines the transformer and its parameters in a configuration file such as in Figure \ref{fig:transformer_params}. Our framework reads in the \textbf{transformer$\_$params} dictionary and creates a transformer object for each entry. In the DGA classification task, as the input object is a single text string representing a domain name, thus we only need a single string transformer. The transformer definition also includes some initialization arguments. The \textbf{subtransformer$\_$args} define the modification actions, which we denote as subtransformers, that can be performed on the data type. In this example, we allow insertion, substitution, and deletion of alphanumeric characters in the string. Note that the modification actions may also have their own initialization arguments (\eg only numerical characters can transformed). The \textbf{input$\_$constraints} define constraints that must be true after an input has been transformed. In this example, we have a simple action constraint that stipulates a string can only be modified three times. Furthermore, there is an additional constraint that insertion and substitution actions can only be performed three times and the deletion action can only remove at most half of the original string. Finally, the \textbf{input$\_$processor$\_$name}, informs our framework that before and after transforming the domain name, a user-defined function of the specified name must be used to process the domain name. This function is one process that ensures semantic correctness and functionality of the transformer input 

\begin{figure}[h] \centering
\includegraphics[width=\linewidth]{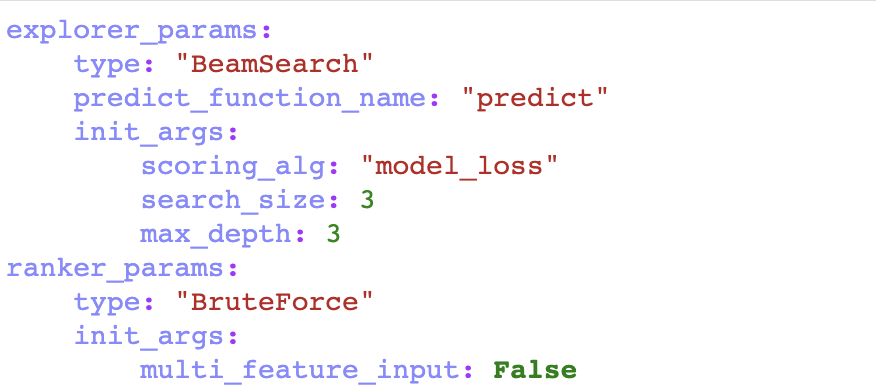}
\caption{The explorer definition for the DGA task.} \label{fig:explorer_params}
\end{figure}

In addition to defining the transformers in the configuration file, the user also defines the explorer parameters as shown in Figure \ref{fig:explorer_params}. Observe that we split the ranking and exploration algorithm parameters. Both parameter sets define the \textbf{type} of algorithm to use as well as the initialization arguments. The vertex scoring algorithm is given in \textbf{scoring$\_$alg} and we also provide the parameters specific to \beam. We note that with respect to the ranking algorithm, an additional Boolean flag, \textbf{multi$\_$feature$\_$input}, is set. It is False by default, but we include it here for illustration. This flag is only True for input types that combine multiple data types and would require multiple transformers such as with tabular data. Finally, due to the numerous machine learning frameworks and model definitions, the \textbf{predict$\_$function$\_$name} is an optional value that informs the framework what function can be used with the model object to obtain a prediction. By default, it is set to ``predict''.

\begin{figure}[h] \centering
\includegraphics[width=\linewidth]{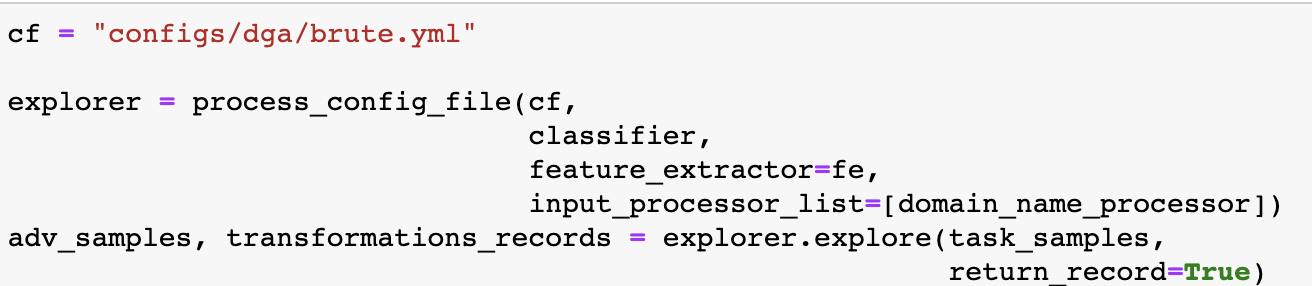}
\caption{Using the explorer to generate samples.} \label{fig:explorer_use}
\end{figure}

After the user loads a model to attack and defines required input processing and feature extraction functions, the configuration file is automatically processed by our framework and an explorer object is returned. Calling it automatically generates adversarial samples for a given set of inputs as shown in Figure \ref{fig:explorer_use}.

%% file: sections/experiments.tex
\section{Evaluation} \label{sec:exp} 
In this section, we present three case studies in which we used our framework as part of an ongoing model evaluation task to generate adversarial inputs. The generic nature of our framework was key in enabling such an evaluation due to differences in input data representations, input semantics, and model types. First, we use the 2018 Home Mortgage Disclosure Act (HMDA) dataset and a mortgage approval classifier to explore \textit{adversarial sample generation for tabular inputs}, an input data type not commonly studied in current adversarial work. Next, we revisit \textit{text and numerical inputs} and generate adversarial samples for the DGA classification task. Finally, we explore generation of adversarial samples for a well-studied, but uncommon input data type, \textit{a binary object}, using a model trained on the EMBER malware classification dataset~\cite{2018arXiv180404637A}. Across all three case studies, we report the following metrics when evaluating our exploration algorithm configurations:
 \begin{enumerate}
     \item Success rate: The number of adversarial inputs that were misclassified by the classifier.
     \item Average number of transforms: The average number of transformations applied to the original input before misclassification occurred. Note this metric is only computed across \textit{successful} adversarial inputs.
     \item Average time per sample: The average runtime of the exploration algorithm to return a potentially adversarial input. This time does not include time required for training for certain algorithms (\ie \lookup and \external).
 \end{enumerate} 
 
 While our framework may not be the most efficient compared with individual prior work tailored for a specific domain, our framework is generic, \ie we can generate adversarial samples for \textit{any input data type}, which enables its use in the evaluation pipeline. Even as we switched between machine learning tasks, the necessary modifications were mostly small-scale changes to our configuration files.

\subsection{Tabular Input}
In this first case study, we use our framework to generate adversarial examples for tabular inputs. To our knowledge, little to no prior works have studied adversarial attacks using tabular inputs or, more generally, inputs that are a combination of multiple data types. The closest works would be adversarial attacks against multi-modal systems~\cite{multimodal1, multimodal2}.

The 2018 release of the Home Mortgage Disclosure Act (HMDA) dataset contains mortgage data from 5,683 institutions. This dataset is a tabular dataset consisting of mix of categorical and numerical features. The classification task is given a tabular data point representing a consumer's mortgage application, predict if the application should be approved or rejected. We were asked to perform an evaluation of the adversarial robustness of five different pre-trained classifiers: a decision tree (DT) classifier, a gradient boosted classifier (GBC), a logistic regression (LR) classifier, a random forest (RF) classifier, and a multi-layer perceptron (MLP) classifier. The accuracy and F1 Score of each classifier are given in Table~\ref{tab:HMDA_performance}. We were also given a pre-processed version of the 2018 HMDA dataset. Compared to the original dataset, the pre-processing sanitized it and, through feature selection, extracts 13 features for classification.

\begin{table}[htb]
\begin{center}
\caption{The performance of each of the HMDA classifiers.}
\label{tab:HMDA_performance}
\resizebox{.9\columnwidth}{!}{%
\begin{tabular}{@{}ccc@{}}
\toprule
\textbf{Model Arch.} & \textbf{Accuracy} & \textbf{F1 Score}\\
\midrule
Decision Tree & 91\% & 0.95\\
Gradient Boosted Classifier & 95\% & 0.97 \\
Logistic Regression & 69\% & 0.81 \\
Random Forest & 81\% & 0.89 \\
Multi-layer Perceptron & 83\% & 0.90 \\
\bottomrule
\end{tabular}
}
\end{center}
\end{table}

\begin{table}[htb]
\begin{center}
\caption{The 7 features selected for adversarial modification from the HMDA dataset}
\label{tab:HMDA_features}
\resizebox{.8\columnwidth}{!}{%
\begin{tabular}{@{}ccc@{}}
\toprule
\textbf{Feature Name} & \textbf{Type} & \textbf{\# of Possible Values}\\
\midrule
Age & Categorical & 7\\
Score Type & Categorical & 8\\
Underwriter & Categorical & 6\\
Loan Limit & Categorical & 2\\
Loan Duration & Categorical & 2\\
Gender & Categorical & 2\\
Race & Categorical & 2\\
\bottomrule
\end{tabular}
}
\end{center}
\end{table}

Based our understanding of task and evaluation goals, we selected 7 categorical features of the 13 total features as potential candidates for modification. We describe these features in Table \ref{tab:HMDA_features}. In our configuration files, we defined 7 categorical transformers. With respect to exploration algorithm specific parameters, the \beam algorithm used a width of 5 and a depth of 2, \ie an input could be transformed at most twice. The \lookup ranking algorithm was trained on 500 randomly selected training inputs and the \external was a reinforcement learning model (See Appendix). The \simanneal algorithm was given a time budget of 1 s per sample and also was limited to 3 transformations. Finally, we use the classification loss scoring function to guide exploration.

In Table~\ref{tab:HMDA_results}, we report the results for the 5 potential exploration algorithm configurations on each of the models we were given using 1000 approved and 1000 rejected applications. As a baseline comparison, we include the results for \beam using the Random algorithm. Within the \beam configurations, the \brute ranking algorithm had the highest success rate across all experiments. It also uses the fewest number of transformations to craft an adversarial sample. The \lookup ranking algorithm is faster that \brute due to the pre-training step, but had a lower success rate as it ignores the current state of the input when determining the next transformation to apply.  We see that the \external ranking algorithm improves upon \lookup with respect to success rate, but as its exploration is still based on an estimation of the edge weights, it uses more transformations to succeed. \simanneal with its fixed time budget also has a high success rate.

% A different version of the table
\begin{table*}[htb]
\centering
\caption{HMDA experimental results.}
\label{tab:HMDA_results}
\begin{tabular}{c|c|ccc} \toprule
\textbf{Model Arch.} & \textbf{Algorithm} & \textbf{Success Rate} & \textbf{Avg. \# of Transforms}  & \textbf{Avg. Time/sample}\\
\midrule
    \multirow{5}{*}{\textbf{Decision Tree}} & \beam (Random) & 38\% & 1.30 & 0.001 s\\
    \cmidrule{2-5}
    & \beam (\brute) & 92\% & 1.13 & 0.010 s\\
    & \beam (\lookup) & 89\% & 1.63 & 0.002 s\\
    & \beam (\external) & 81\% & 1.85 & 0.018 s\\
    & \simanneal & 97\% & 1.87 & 1.000 s\\
\midrule
    \multirow{5}{*}{\textbf{Gradient Boosted Classifier}} & \beam (Random) & 14\% & 1.43 & 0.003 s\\
    \cmidrule{2-5}
    & \beam (\brute) & 58\% & 1.08 & 0.044 s\\
    & \beam (\lookup) & 26\% & 1.41 & 0.026 s \\
    & \beam (\external) & 52\% & 1.74 & 0.058 s\\
    & \simanneal & 57\% & 2.00 & 1.000 s\\
\midrule
    \multirow{5}{*}{\textbf{Logistic Regression}} & \beam (Random) & 34\% & 1.38 & 0.002 s\\
    \cmidrule{2-5}
    & \beam (\brute) & 100\% & 1.05 & 0.007 s\\
    & \beam (\lookup) & 69\% & 1.12 & 0.007 s\\
    & \beam (\external) & 88\% & 1.93 & 0.020 s\\
    & \simanneal  & 100\% & 2.00 & 1.000 s\\
\midrule
    \multirow{5}{*}{\textbf{Random Forest}} & \beam (Random) & 27\% & 1.46 & 0.352 s\\
    \cmidrule{2-5}
    & \beam (\brute) & 100\% & 1.04 & 1.462 s\\
    & \beam (\lookup) & 70\% & 1.08 & 1.177 s\\
    & \beam (\external) & 86\% & 1.96 & 0.042 s\\
    & \simanneal  & 75\% & 1.87 & 1.000 s\\
\midrule
    \multirow{5}{*}{\textbf{Multi-Layer Perceptron}} & \beam (Random) & 36\% & 1.41 & 0.198 s\\
    \cmidrule{2-5}
    & \beam (\brute) & 100\% & 1.04 & 0.724 s\\
    & \beam (\lookup) & 94\% & 1.39 & 0.369 s\\
    & \beam (\external) & 71\% & 1.92 & 0.297 s\\
    & \simanneal  & 97\% & 1.90 & 1.000 s\\
\bottomrule
\end{tabular}
\end{table*}

\subsection{Text Input}
\label{sec:dga_exp}

In our second case study, we use our framework to generate adversarial examples for text inputs. We recognize that much prior work exists with respect to generating adversarial text inputs, namely Counterfit~\cite{counterfit} and TextAttack~\cite{ morris2020textattack}. While these works may be more efficient or successful at finding adversarial text inputs compared to our framework, they cannot be easily adapted to other input data types.

Domain name generation algorithms (DGA) are often used by malware to locate the command and control servers and avoid simple blacklisting. Instead of relying on fixed IP address or domain name hard-coded into the malware, domain name generation algorithms generate a large of number of domain name strings in a pseudorandom fashion, one of which connects to the actual command and control server~\cite{dga}. The classification task is given a domain name, predict if was produced by a domain name generation algorithm. We were provided with a DGA classifier that was a 2-layer neural network. We were also given a test dataset containing 5,000 DGA names and 5,000 non-DGA names. Given a domain name, the classifier extracts 20 numerical bounded features to use for classification. The accuracy of the classifier on the test dataset is 93\%.

As the domain name is a single string, we defined a string transformer with the ability to add, delete, or substitute alphanumeric characters in the string. We also defined a function to preprocess the domain name and extract its top level domain (\eg ``.com'') as this portion of the string was not valid for modification. With respect to exploration algorithm specific parameters, the \beam algorithm used a width of 3 and a depth of 3, \ie an input could be transformed at most three times. The \lookup ranking algorithm was trained on 500 randomly selected training inputs and the \external algorithm was trained on all of the samples in the test set. The \simanneal algorithm was given a time budget of 1 s per sample and was also limited to 3 transformations. Finally, we use the classification loss scoring function to guide exploration.

In Table \ref{tab:dga:pipeline}, we report the results of our study on 1000 DGA and 1000 non-DGA samples. As a baseline comparison, we include the results for \beam using the Random algorithm. As before, we observe that the \brute algorithm has the highest success rate and the fewest number of transformations. The \lookup algorithm is faster in comparison, but uses more transformation. The \external algorithm, in this case, has worse performance compared to \brute and requires more transformations on average. Finally, \simanneal also has a high success rate and allows the user to exactly specify exploration time, but usually maximizes its transformation budget.

\begin{table}[htb]
\centering
\caption{DGA experimental result - Generating Adversarial Objects with Classifier Loss Objective.}
\label{tab:dga:pipeline}
\resizebox{\columnwidth}{!}{%
\begin{tabular}{c|ccccc} \toprule
    Algorithm & Success rate & Avg. \# of Transforms & Avg. Time / sample \\ \midrule
    \beam (Random) & 23\% & 1.84 & 0.093 s \\
    \midrule
    \beam (\brute) & 85\% & 1.24 & 0.363 s \\
    \beam (\lookup) & 45\% & 1.61 & 0.277 s \\
    \beam (\external) & 70\% & 2.56 & 0.400 s \\
    \simanneal & 62\% & 2.28 & 1.000 s \\
    \bottomrule
\end{tabular}
}
\end{table}

 As the classifier's feature extractor generates an array of continuous numerical features, we can use existing image based adversarial attacks to generate an adversarial feature vector for a given domain name. In our second evaluation, we used the Projected Gradient Descent (PGD) attack~\cite{madry2018towards} to generate adversarial feature target vectors for the 1000 DGA and 1000 non-DGA samples. Then, we define the feature distance scoring function to be the cosine similarity between the current feature perturbation induced by an input transformation and the PGD target vector.  We also set a modification constraint specifying that a feature can be only modified by up to 30\% of the feature's maximum value in order to mimic the inconspicuousness property image-based adversarial attacks enforce. Except for the change in scoring function, we use the same exploration parameters as before.
 
 In Table \ref{tab:dga:fe}, we report the results of this evaluation. Of note is that the \external ranking algorithm is much faster compared to the other \beam configurations. Clearly, the RL model is very effective at estimating the current and future effects a particular transformation induces, leading to an overall decrease in time required to generate an adversarial sample. However, we see there is an overall drop in success rate and an increase in both average number of transformations and average time per sample for the other exploration algorithm configurations. The drop in success rate is likely due to the fact that some of the adversarial target vectors generated by the PGD attack are not realizable. The increase in number of transformation and time per sample indicates that the selected input transformations, although resulting in a feature perturbation in the target direction, are likely not the most efficient path to generate a misclassified adversarial input. This particular experiment shows that our framework is backwards compatible with traditional adversarial attacks through use of the feature distance scoring function.

\begin{table}[htb]
\centering
\caption{DGA experimental result - Generating Adversarial Objects with Feature Distance Objective.}
\label{tab:dga:fe}
\resizebox{\columnwidth}{!}{%
\begin{tabular}{c|ccccc} \toprule
    Algorithm & Success rate & Avg. \# of Transforms & Avg. Time / sample \\ \midrule
    \beam (Random) & 27\% & 1.87 & 0.091 s \\
    \midrule
    \beam (\brute) & 56\% & 1.93 & 22.835 s \\
    \beam (\lookup) & 50\% & 1.79 & 12.415 s \\
    \beam (\external) & 43\% & 2.69 & 0.606 s \\
    \simanneal & 26\% & 2.72 & 1.000 s \\
    \bottomrule
\end{tabular}
}
\end{table}

As our framework enables the generation of adversarial inputs regardless of the task or input type, we now have a method to improve the model’s accuracy on adversarial samples. Adversarial training is a state-of-the-art defense that improves a model's adversarial accuracy by generating adversarial inputs on-the-fly during training.~\cite{madry2018towards}. As we showed in Table~\ref{tab:intro:at}, although adversarial training could be used with adversarial features, it was not very effective on real adversarial objects. In Table~\ref{tab:dga:at}, we report the performance of the DGA classifier when trained using standard training and adversarial training. As we see, a model trained on adversarial objects exhibits much higher adversarial accuracy with minimal impact to it natural (\ie non-adversarial) accuracy compared to training with adversarial features.

\begin{table}[htb]
    \centering
    \caption{Evaluation of the adversarial accuracy of the DGA model when trained using standard and adversarial training.  Standard training denotes training on the original unmodified dataset. Standard adversarial training uses PGD to generate adversarial feature vectors. DGA Adversarial training uses adversarial objects generated by \beam (\external). Note that the ranking model was trained on a pre-trained DGA classifier trained on the original dataset. }
    \label{tab:dga:at}
    \resizebox{\columnwidth}{!}{%
    \begin{tabular}{c|cc} \toprule
    Training Method & Natural Accuracy & Adversarial Accuracy \\ 
        \midrule
        Standard training & 92\% & 19\% \\
        Standard Adversarial Training (Adversarial Features) & 82\% & 29\% \\
        \midrule
        Adversarial Training with Adversarial Objects (Ours) & 92\% & \textbf{45\%} \\ \bottomrule
    \end{tabular}
    }
\end{table}

\subsection{Non-Standard Input - Binary}

In our last case study, we use our framework to adversarial examples for a non-standard input type, binary files. With respect to manipulated binary inputs, there are several works that proposed adversarial attack algorithms against malware classifiers~\cite{Demetrio2020AdversarialEA, grosse2017adversarial, anderson2017, 10.1145/3433210.3453086}, but, as with text inputs, these algorithms are not generic. It is a non-trivial task to adapt their proposed methods to other input data types. 

We also focus our study on the malware classification task using the EMBER dataset, a collection of feature vectors extracted from 1.1 million binary files with goodware/malware labels~\cite{2018arXiv180404637A}. Given a binary file, the classifier must predict if the file is malware or goodware. We use a pre-trained classifier trained on 900,000 training data points in which one third of the training data is unlabeled. The remaining two thirds are labeled and balanced between malware and goodware. On the 200,000 test data points, the accuracy of the classifier is 97\%. The model authors also provide a binary file feature extractor and SHA256 hashes of the dataset binaries so that the original binary files can be collected.

Unlike our previous case studies, binary files are a non-standard input type so we must first define a new transformer class responsible for perturbing and ensuring semantic correctness and functionality of the transformer binary. Drawing from prior work~\cite{anderson2018learning}, we implemented six types of functionality-preserving transformations which modify the PE header: binary (un)packing, adding sections, renaming sections, adding imports, removing debugging header information, and appending header data. In this study, we will only report results for a \beam exploration with a \external ranking algorithm. The ranking model follows reinforcement learning method described in the Appendix\footnote{As malicious binaries are risky to handle, we train the algorithm multiple times on 12 benign binaries.}. We also use the feature distance scoring objective, but instead use the $L_2$ distance between the current perturbation vector and target perturbation vector to guide exploration. The target adversarial perturbation vectors are generated using the HopSkipJump algorithm~\cite{HSJ} against the previously described malware classifier. We evaluate the malware classifier against 93 correctly labelled malware samples from the test dataset. As the original dataset only contains the extracted feature vectors, we use the provided SHA256 hashes to obtain the original malware binaries. With a beam width of 5 and a depth of 3, \ie a binary could be transformed at most 3 times, our framework generated 38 adversarial malware binaries. 

It is a well-known fact that adversarial examples in the image domain are transferable~\cite{szegedy2014intriguing}, \ie adversarial examples generated for one model are highly likely to be misclassified by other models trained for the same task. In Table \ref{tab:vt}, we report the success rate of the previously generated 38 adversarial malware binaries against a set of VirusTotal detectors. We see that the adversarial modifications performed by our framework cause the average detection rate to fall from 82\% to 58\%. In the worst case, one of the detectors only flagged 6\% of the binaries are malicious. In Table~\ref{tab:vtind}, we provided the  precision and recall of 6 different anonymized detectors found in VirusTotal computed across the original and adversarial binaries. While we recognize that VirusTotal is a mix of signature and non-signature based detectors, it remains true that many organizations rely on VirusTotal to label malicious binaries. These results suggest that adversaries can exploit adversarial transferability to evade detection, especially if more advanced input transformations are used.

\begin{table}[htb]
\centering
\caption{The detection rate of the VirusTotal detectors on the original and adversarially modified malware binaries.}
\label{tab:vt}
\resizebox{.7\columnwidth}{!}{%
\begin{tabular}{c|ccccc} \toprule
    Dataset & Average & Min  & Max \\ \midrule
    Original Binaries & 82\% & 50\% & 90\%  \\
    Adversarial Binaries & 58\% & 6\% & 91\%  \\
    \bottomrule
\end{tabular}
}
\end{table}

\begin{table}[htb]
\centering
\caption{Performance of 6 different VirusTotal Detectors.}
\label{tab:vtind}
\resizebox{.9\columnwidth}{!}{
\begin{tabular}{c|cc|cc} \toprule
            & \multicolumn{2}{|c|}{Precision}
            & \multicolumn{2}{|c}{Recall} \\ \midrule
    Scanner & Original & Adversarial & Original & Adversarial \\ \midrule
    A & 100\% & 76\% & 100\% & 91\% \\
    B & 100\% & 68\% & 100\% & 56\% \\
    C & 100\% & 77\% & 100\% & 81\% \\
    D & 100\% & 100\% & 99\% & 70\% \\
    E & 100\% & 60\% & 60\% & 58\% \\
    F & 100\% & 100\% & 43\% & 28\% \\
    \bottomrule
\end{tabular}
}

\end{table}

%% file: sections/conclusion.tex
\section{Conclusion} \label{sec:conclusion}
The vulnerability of AI systems to adversarial machine learning motivates a need for tools that enable proper study. As traditional adversarial attack algorithms were designed for image-based systems, strict assumptions regarding input structure have been made that limit the use of such algorithms in other data domains. In this paper, we proposed a new adversarial attack framework capable of generating adversarial inputs for AI systems regardless of the modality or task. We represent the adversarial generation process as a graph exploration problem. Our framework searches for a sequence of edges representing input transformations that result in satisfying the exploration object, \eg finding an adversarial input. We pre-define several data transformers for common input types. In addition, we provide a transformation interface to address cases where a particular data type is unsupported. In such cases, the user can rely on their domain knowledge to define the new data transformer, which can be integrated into our tools for future use, as we did when defining binary transformations. In an effort to ease usability, we allow users to customize the adversarial generation process through use of configuration files, abstracting away many of the low level implementation details. As we highlight in our three evaluation case studies, the current version of our framework contains several possible exploration configurations, with respectable effectiveness despite the variety in data representation, model types, and semantic/functionality constraints. Switching between machine learning tasks in the studies only required small scale modifications to the exploration parameters and data transformation definitions.

As we are making our toolkit publicly available\footnote{https://github.com/IBM/URET}, we recognize there are ethical concerns regarding malicious use of our tools, motivating a push to privatize our framework as Pierazzi \etal have done~\cite{pierazzi2020problemspace}. However, we do not believe that keeping such tools private will benefit the long term health of machine learning as  malicious actors can still develop such tools independently. Rather, by providing these tools to the community, we provide academic and industry researchers alike the ability to study the effects of adversarial attacks in new input domains and develop generic adversarial mitigation techniques.

\paratitle{Acknowledgements} We thank Mohinder Singh for providing us with the pre-trained HMDA models and our anonymous reviewers for their valuable feedback. This research was developed with funding from the Defense Advanced Research Projects Agency (DARPA). The views, opinions and/or findings expressed are those of the author and should not be interpreted as representing the official views or policies of the Department of Defense or the U.S. Government. Distribution Statement “A” (Approved for Public Release, Distribution Unlimited). 

%% file: sections/appendix.tex
\appendix

\section{Model used for \external Ranking}
\label{sec:app_rl}

There are multiple ways to train a model to map an input to select the best action (\ie input transformation) to produce an adversarial example. In our experiments, we use reinforcement learning for the \external algorithm because such models consider the effect a transformation has on both current and future rewards.  In our specific implementation, we adapted Deep Q Learning~\cite{deepq}, to our problem domain.

In general, Deep Q Learning trains a neural network or \emph{neural policy} that takes the current state to predict the rewards of actions.
The training process samples an action from the neural policy and applies it to get actual reward score. If it leads to a state with a higher reward, the model is updated to take more of the chosen action for the similar states.
The reward for an action includes the future reward to backpropagate the effect to earlier action choices.
In our case, the reward is vertex score, and actions are input transformations, and we use this Deep Q Learning implementation provided by Keras-RL~\cite{plappert2016kerasrl} for this process.

Our main adaptation is on training policy that decides which action to take during training.
Plain Deep Q Learning using a randomly initialized neural policy can have 
low chance of success due to the high complexity of search space.
The large number of possible edges and edge combinations in our graph makes learning an effective neural policy difficult as Deep Q Learning is normally trained with a small action space (\eg 2-16 actions). To address this, rather than
choosing a random sample as the starting vertex and letting the model explore the entire graph with a randomly initialized policy, we do the following for each training:

\begin{enumerate}
    \item Pre-generate a goal vertex using a sequence of random transformations on the current training sample. This sequence of transformations represents the \emph{ideal sequence} of transformations for the current training input. %with which to guide training for this epoch.
    \item At each transformation step, randomly generate an action from one of three sources:
        \begin{itemize}
            \item The ideal sequence - This is the pre-generated transformation sequence that acts as an answer key
            \item The current neural policy - This is the policy learned by the reinforcement learning model.
            \item The Random policy - This simply selects one of the available transformations based on the current input state.
        \end{itemize}
    \item Once an end state has been reached, use the exploration object (\eg classifier loss) and the ideal sequence to evaluate the fitness of the selected actions and update the neural policy.
\end{enumerate}

The above process repeats for all inputs in the training set and can be looped multiple times. Furthermore, as training progresses, the actions generated in step 2 progressively become more likely to be from the current neural policy rather than the ideal sequence of the Random policy.

\section{Numerical Inputs}

As the DGA classifier first extracts 20 numerical continuous features, we also tested our framework on numerical data. Specifically, we were interested in the success rate of our framework at generating adversarial feature vectors, similar to traditional adversarial attacks, but with some constraints on the transformation process. We selected 13 of the 20 numerical features that, when modified, were likely to be realizable if necessary (\eg relative number of consonants/vowels, number of dashes, length, etc.). Each transformer was only allowed to modify at most 30\% of the feature's current value. As one of the features recorded the length of the domain name, we added an additional constraint to this transformer that the length could only be increased if modified.
We present these results in Table~\ref{tab:dga:model}. We observe an overall increase in success rate and decrease in average time per sample compared to Table \ref{tab:dga:fe}. Note that we did not evaluate the \external algorithm for this experiment.

\begin{table}[htb]
\centering
\caption{DGA experimental result - Generating Adversarial Features.}
\resizebox{\columnwidth}{!}{%
\label{tab:dga:model}
\begin{tabular}{c|ccccc} \toprule
    Algorithm & Success rate & Avg. \# of Transforms & Avg. Time / sample \\ \midrule
    \beam (Random) & 6\% & 1.87 & 0.010 s \\
    \midrule
    \beam (\brute) & 65\% & 1.73 & 0.801 s \\
    \beam (\lookup)  & 42\% & 2.02 & 0.069 s \\
    \simanneal & 52\% & 2.96 & 1.000 s \\
    \bottomrule
\end{tabular}
}
\end{table}

%%%%%%%%%%%%%%%%%%%%%%%%%%%%%%%%%%%%%%%%%%%%%%%%%%%%%%%%%%%%%%%%%%%%%
% Artifact Appendix
%%%%%%%%%%%%%%%%%%%%%%%%%%%%%%%%%%%%%%%%%%%%%%%%%%%%%%%%%%%%%%%%%%%%%
\newcommand{\artifacturl}[0]{\url{https://github.com/IBM/URET/tree/8bd1b4f4d78ac19f026e862b31ae933983c99551}}

\section{Artifact Appendix}

\subsection{Abstract}
The provided artifact contains URET as described in the paper. The tools provided are sufficient to allow users to perform custom adversarial evaluations on machine learning classifiers regardless of input domain. Specifically, this version includes input transformer definitions for the common input types (e.g., numerical, text, and categorical) as well as the binary file input type described in the paper. With respect to results reproduction, it contains the model checkpoints, evaluation data, and notebooks used to generate most of the results in Table 6.

Some components described in the paper have been purposely left out of the provided artifact for proprietary reasons:
\begin{itemize}
    \item No implementation of the ``Model Guided'' algorithm. This implementation was deemed proprietary. 
    \item No data/notebooks for the Malware experiments. The malware samples used for evaluation are proprietary and may pose a risk if improperly handled. 
    \item No data/notebooks for the DGA experiments. The training and evaluation data are proprietary. The model code is also proprietary.
\end{itemize}

Despite these limitations, the provided artifact can be used to perform the custom evaluations described in the paper.

%%%%%%%%%%%%%%%%%%%%%%%%%%%%%%%%%%%%%%%%%%%%%%%%%%%%%%%%%%%%%%%%%%%%%
\subsection{Description \& Requirements}

\subsubsection{Security, privacy, and ethical concerns}
There should be no security, privacy, and ethical concerns.

\subsubsection{How to access}

URET is an evaluation toolkit for adversarial evasion attacks. The public URET repository is accessible at \url{https://github.com/IBM/URET}. It should contain the code necessary to perform an evaluation as well as notebook examples of how to use URET. The stable URL used for Artifact Evaluation is \artifacturl.

\subsubsection{Hardware dependencies}
Our artifact does not have any hardware requirement. Of note, a GPU is not required to run the evaluation notebooks and pre-trained model checkpoints have been provided. We tested our artifact on an Ubuntu 18.04 system with 8 CPU cores.

\subsubsection{Software dependencies}

The artifact repository contains a setup script for installing the required python libraries to use URET, independent of any machine learning libraries (e.g., Tensorflow, PyTorch, etc.). However, the example notebooks require a different setup script, which is included in the artifact, as the model checkpoints were trained using older libraries. The example notebooks were tested using Python 3.8. In Section A.3., we detail the necessary steps to install the required python libraries. We recommend evaluators create a virtual environment prior to running the setup script.

\subsubsection{Benchmarks}
The artifact requires the 2018 Home Mortgage Disclosure Act (HMDA) dataset to run the evaluation notebooks. We have already included a copy of the dataset in the artifact.

%%%%%%%%%%%%%%%%%%%%%%%%%%%%%%%%%%%%%%%%%%%%%%%%%%%%%%%%%%%%%%%%%%%%%
\subsection{Set-up}

\subsubsection{Installation}

Here, we provide instructions to deploy URET and run the evaluation notebooks. This was tested using Python 3.8.

\begin{enumerate}
    \item Clone the artifact from the stable URL: \artifacturl
    \item In the artifact directory, replace the existing \hl{setup.py} file with the version from from \hl{notebooks/setup.py}.
    \item Run the setup script in the top level directory (i.e. \hl{pip install -e .}) to install the evaluation libraries. It is suggested you do this in a virtual environment.
\end{enumerate}

After step 3, URET should be ready for use. To reproduce most of the results in Table 6, move in to the \hl{notebooks/} directory and run \hl{HMDA\_results.ipynb}. We have included pre-computed adversarial examples generated from running each of the exploration algorithms described in the paper. This samples are stored in \hl{notebooks/data/HMDA\_adv\_samples}. Note that running a generation notebook (e.g., \hl{notebooks/HMDA\_random.ipynb}) will overwrite the saved samples by default.

\subsubsection{Basic Test}
After setting up URET, the easiest method to test functionality is to run one of the adversarial generation notebooks. We recommend running \hl{notebooks/HMDA\_random.ipynb} as it is the fastest algorithm to run. The notebook should run without errors, though you may get some warning messages. Cell 4 should display several progress bars and text related to the model being evaluated and the adversarial success rate of the generated samples.

%%%%%%%%%%%%%%%%%%%%%%%%%%%%%%%%%%%%%%%%%%%%%%%%%%%%%%%%%%%%%%%%%%%%%
\subsection{Evaluation workflow}

\subsubsection{Major Claims}

\begin{compactdesc}

    \item[(C1):] \textit{URET can be used to perform generating adversarial evasion examples for a variety of input domains and formats. This claim proven by experiment E1 and Sections 6.2 and 6.3 in the paper. Experiment E1 is described in Section 6.1 of the main paper and its results are reported in Table 6.}
    
    \item[(C2):] \textit{URET can generate adversarial examples for inputs containing a multiple features in the input. This claim proven by experiment E1.}

    \item[(C3):] \textit{URET can generate adversarial examples for inputs containing a single feature. This claim proven by Sections 6.2 and 6.3 in the paper.}
    
    \item[(C4):] \textit{URET presents several different exploration confirmations that can be selected based on user needs. Experiment E1 shows results for several exploration configurations as well as a baseline (Random) to highlight the success rate/speed tradeoff.}

\end{compactdesc}

\subsubsection{Experiments}

% use paralist for more compact list format: for more details check here:
% https://texfaq.org/FAQ-complist
\begin{compactdesc}

    \item[(E1):] \textit{[HMDA Adversarial Examples][6 Human-hours]:
   Generate adversarial examples for five HMDA classification models using four exploration configurations.}

    \begin{itemize}[noitemsep,nolistsep,wide=0pt]
        \item[\textbf{Preparation:}] Follow the installation instructions in Section A.3.1. Once URET has been installed along with its required libraries, change to the \hl{notebooks/} directory before beginning the experiment.

        \item[\textbf{Execution:}]
        We have pre-computed adversarial examples for each of the generation notebooks. If the evaluator wants to re-generate the adversarial examples, then run the following notebooks:
        \begin{enumerate}
            \item \hl{notebooks/HMDA\_random.ipynb} - [~25 Human-minutes] This generates adversarial examples where every transformation edge is selected randomly.
            \item \hl{notebooks/HMDA\_brute.ipynb} - [~1.5-2 Human-hours] This generates adversarial examples by exploring every edge.
            \item \hl{notebooks/HMDA\_lookup.ipynb} - [~1 Human-hour] This generates adversarial examples using the lookup table algorithm. Each generation process will first compute a transformation weight lookup table followed by generation.
            \item \hl{notebooks/HMDA\_simanneal.ipynb} - [2.5-3 Human-hours] This generates adversarial examples using the simulated annealing algorithm. The default configuration file assigns ~1 sec of attack time per sample.
        \end{enumerate}
        
        Of the models, the random forest and multi-layer perception models require the most amount of time to generate.

        \item[\textbf{Results:}] To output attack success rate and transformation count on generated adversarial examples, run \hl{notebooks/HMDA\_results.ipynb}. It expects adversarial examples for each exploration configuration and model. 
        
        To get the per sample generation times, we divide the generation time shown in the generation notebooks by the number of samples (2000). We have noticed that the simulated annealing generation time can be sometimes longer than the specified amount. 
    \end{itemize}

\end{compactdesc}

We have provide configuration files for each of the experiments shown in Table 6 of the paper. Due to randomness, there may be some slight variation in the success rate, transformation count, and per sample generation time between adversarial example generations. The configuration files can be found in \hl{notebooks/configs/HMDA}. If interested, the evaluator can alter these configuration files to try different exploration settings. For the non-simulated annealing configuration files, consider modifying the beam width (i.e., how many potential transformation candidates are considered) and beam depth (i.e., how many transformations can be applied). For simulated annealing, consider modifying the transformation parameters:
\begin{itemize}
    \item max\_transform\_i\_sampled - Upper limit on feature transformation applied in a single transformation step
    \item global\_max\_transforms - how many transformations can be applied
\end{itemize}
or the attack time. We note that for simulated annealing, modifying the number of transformations without increasing attack time may result in decreasing the success rate given its random exploration process.

Note that we do not include the code, models, or data for the experiments show in sections 6.2 and 6.3 in main paper. We are unable to share the relevant material due to the proprietary nature of data. Section 6.1 is the only experiment that uses entirely non-proprietary data.

%%%%%%%%%%%%%%%%%%%%%%%%%%%%%%%%%%%%%%%%%%%%%%%%%%%%%%%%%%%%%%%%%%%%%
\subsection{Notes on Reusability}
\label{sec:reuse}
URET is intended to be an evolving set of tools that can be used to evaluate adversarial robustness of classifiers with respect to evasion. If users find the current set of modules insufficient for their needs, they are encouraged to implement their own custom modules using the common interfaces exposed by URET. Specifically, we expect users may need to customize some or all of the following component:
\begin{itemize}
    \item Input Transformers and Subtransformers (found in \hl{uret/transformers}) - For data types beyond the basic and binary types we include in URET, users will need to provide new implementations, which the exploration algorithms can use.
    \item Custom Loss Functions (found in \hl{uret/transformers}) - URET uses two common loss types: 1) classification loss based on ground truth labels or model predictions and 2) Distance based loss function. Users that require alerted or unique loss functions (e.g., a loss based on time series input data) can define their own function to provide to the explorer during initialization.
    \item Dependencies - Some feature relationships need to be handled outside of the transformation interface, such as normalization of a multi-feature vector input. These dependencies can be functionally defined and specified in the URET configuration file for the explorer to enforce during example generation.
\end{itemize}

The goal of URET was to provide a basic, but easily expandable set of tools to be used for adversarial evaluations. We hope that as users customize URET for their own needs, their implementations can be integrated into the public repository to expand URET's capabilities and help other users.

%%%%%%%%%%%%%%%%%%%%%%%%%%%%%%%%%%%%%%%%%%%%%%%%%%%%%%%%%%%%%%%%%%%%%

\subsection{Version}
%%%%%%%%%%%%%%%%%%%%
% Obligatory.
% Do not change/remove.
%%%%%%%%%%%%%%%%%%%%
Based on the LaTeX template for Artifact Evaluation V20220926. Submission,
reviewing and badging methodology followed for the evaluation of this artifact
can be found at \url{https://secartifacts.github.io/usenixsec2023/}.